\documentclass[10pt,twocolumn,letterpaper]{article}

\usepackage{iccv}
\usepackage{times}
\usepackage{epsfig}
\usepackage{graphicx}
\usepackage{amsmath}
\usepackage{amssymb}
\usepackage[ruled]{algorithm}
\usepackage{enumerate}
\usepackage{algpseudocode}
\usepackage{subfigure}
\usepackage{booktabs}
\usepackage{multirow}
\usepackage{tikz}
\usepackage{pgfplots}
\usepackage{bbm}

\usepackage[pagebackref=true,breaklinks=true,letterpaper=true,colorlinks,bookmarks=false]{hyperref}

\iccvfinalcopy

\def\iccvPaperID{307}

\DeclareMathOperator*{\argmax}{arg\,max}
\DeclareMathOperator*{\CNN}{CNN}
\DeclareMathOperator*{\LSTM}{LSTM}

\ificcvfinal\fi
\begin{document}

\title{Show, Ask, Attend, and Answer: \\
A Strong Baseline For Visual Question Answering}


\author{Vahid Kazemi \qquad Ali Elqursh\\
Google Research\\
1600 Amphitheater Parkway\\
{\tt\small \{vahid, elqursh\}@google.com}
}

\maketitle

\begin{abstract}
This paper presents a new baseline for visual question answering task. Given an image and a question in natural language, our model produces accurate answers according to the content of the image. Our model, while being architecturally simple and relatively small in terms of trainable parameters, sets a new state of the art on both unbalanced and balanced VQA benchmark. On VQA 1.0 \cite{Antol2015VQAVQ} open ended challenge, our model achieves 64.6\% accuracy on the test-standard set without using additional data, an improvement of 0.4\% over state of the art, and on newly released VQA 2.0 \cite{Goyal2016MakingTV}, our model scores 59.7\% on validation set outperforming best previously reported results by 0.5\%. The results presented in this paper are especially interesting because very similar models have been tried before \cite{Yang2016StackedAN} but significantly lower performance were reported. In light of the new results we hope to see more meaningful research on visual question answering in the future.
\end{abstract}

\section{Introduction}
\begin{figure}
\subfigure{\includegraphics[width=150pt]{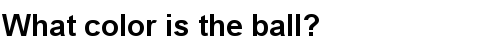}}
\subfigure{\includegraphics[width=110pt]{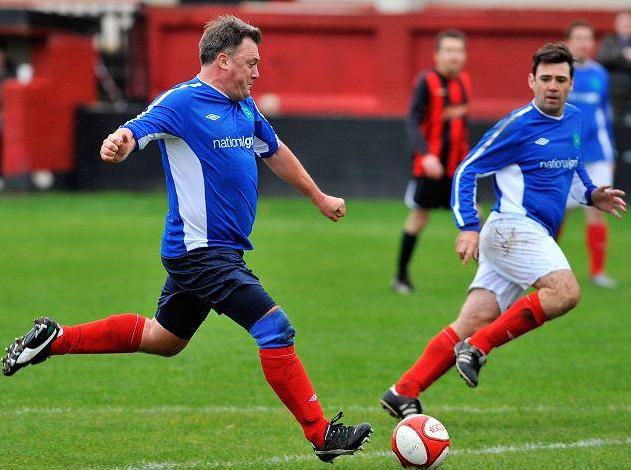}}
\subfigure{\includegraphics[width=110pt]{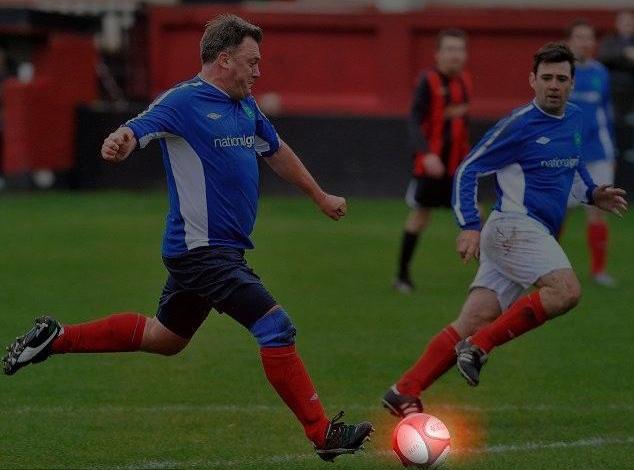}}
\subfigure{\includegraphics[width=90pt]{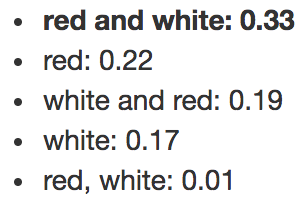}}
\caption{Top 5 predictions from our model and their probabilities for an example image/question pair. On the right we visualize the corresponding attention distribution produced by the model.}
\label{fig:teaser}
\end{figure}
Deep neural networks in the last few years have made dramatic impact in computer vision and natural language processing fields. We are now able to build models that recognize objects in the images with high accuracy \cite{Krizhevsky2012ImageNetCW,Szegedy2015GoingDW,He2016DeepRL}. But we are still far from human level understanding of images. When we as humans look at images we don't just see objects but we also understand how objects interact and we can tell their state and properties. Visual question answering (VQA) \cite{Antol2015VQAVQ} is particularly interesting because it allows us to understand what our models truly see. We present the model with an image and a question in the form of natural language and the model generates an answer again in the form of natural language. 

A related and more throughly researched task to VQA is image caption generation \cite{Xu2015ShowAA,Vinyals2015ShowAT}, where the task is to generate a representative description of an image in natural language. A clear advantage of VQA over caption generation is that evaluating a VQA model is much easier. There is not a unique caption that can describe an image. Moreover, it is rather easy to come up with a single caption that more or less holds for a large collection of images. There is no way to tell what the model actually understands from the image based on a generic caption. Some previous work have been published that tried to mitigate this problem by providing dense \cite{Johnson2016DenseCapFC} or unambiguous captions \cite{Mao2016GenerationAC}, but this problem is inherently less severe with VQA task. It is always possible to ask very narrow questions forcing the model to give a specific answer. For these reasons we believe VQA is a good proxy task for creating rich representations for modeling language and vision.

\begin{figure*}
	\centering
	\includegraphics[width=\linewidth]{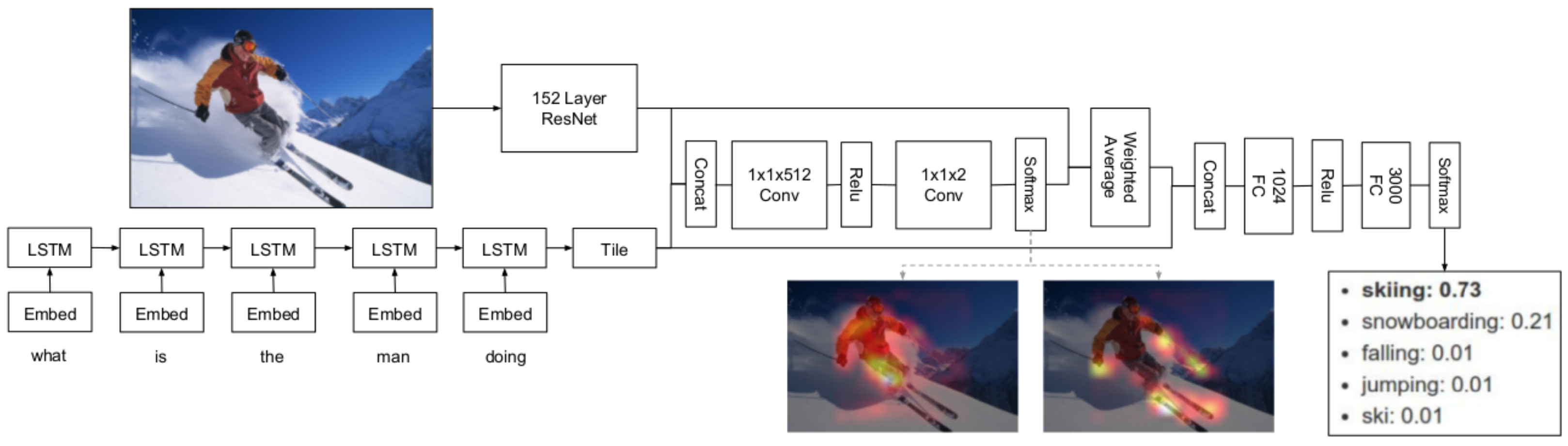}
	\caption{An overview of our model. We use a convolutional neural network based on ResNet \cite{He2016DeepRL} to embed the image.  The input question is tokenized and embedded and fed to a multi-layer LSTM. The concatenated image features and the final state of LSTMs are then used to compute multiple attention distributions over image features. The concatenated image feature glimpses and the state of the LSTM is fed to two fully connected layers two produce probabilities over answer classes. }
	\label{fig:overview}
\end{figure*}

Some novel and interesting approaches \cite{Fukui2016MultimodalCB,Nam2016DualAN}  have been published in the last few years on visual question answering that showed promising results. However, in this work, we show that a relatively simple architecture (compared to the recent works) when trained carefully bests state the art. Figure \ref{fig:overview} provides a high level overview of our model. To summarize, our proposed model uses long short-term memory units (LSTM) \cite{Hochreiter1997LongSM} to encode the question, and a deep residual network \cite{He2016DeepRL} to compute the image features. A soft attention mechanism similar to \cite{Xu2015ShowAA} is utilized to compute multiple glimpses of image features based on the state of the LSTM. A classifier than takes the image feature glimpses and the final state of the LSTM as input to produce probabilities over a fixed set of most frequent answers. On VQA 1.0 \cite{Antol2015VQAVQ} open ended challenge, our model achieves 64.6\% accuracy on the test-standard set without using additional data, an improvement of 0.4\% over state of the art, and on newly released VQA 2.0 \cite{Goyal2016MakingTV}, our model scores 59.7\% on validation set outperforming best reported results by 0.5\%.

This paper proves once again that when it comes to training neural networks the devil is in the details \cite{Chatfield2014ReturnOT}.


\section{Related work}
In this section we provide an overview of related work.

Convolutional neural networks (CNNs) \cite{LeCun2010ConvolutionalNA} have revolutionalized the field of computer vision in the recent years. Landmark paper by Krizhevsky et al. \cite{Krizhevsky2012ImageNetCW} for the first time showed great success on applying a deep CNN on large scale ImageNet \cite{Deng2009ImageNetAL} dataset achieving a dramatic improvement over state of the art methods that used hand designed features. In the recent years researchers have been hard at work training deeper \cite{Szegedy2015GoingDW}, very deep \cite{Szegedy2016RethinkingTI}, and even deeper \cite{He2016DeepRL} neural networks. While success of neural networks are commonly attributed to larger datasets and more compute power, there are a lot of details that we know and consider now that were not known just a few years ago. These include choice of activation function \cite{Nair2010RectifiedLU}, initialization \cite{Glorot2010UnderstandingTD}, optimizer \cite{Kingma2014AdamAM}, and regularization \cite{Hinton2012ImprovingNN}. As we show in this paper at times getting the details right is more important than the actual architecture.

When it comes to design of deep neural networks, very few ideas have been consistently found advantageous across different domains. One of these ideas is notion of attention \cite{Mnih2014RecurrentMO,Vinyals2015ShowAT}, which enables deep neural networks to extract localized features from input data.

Another neural network model that we take advantage of in this work is Long Short-Term Memory (LSTM) \cite{Hochreiter1997LongSM}. LSTMs have been widely adopted by machine learning researchers in the recent years and have shown oustanding results on a wide range of problems from machine translation \cite{Bahdanau2014NeuralMT} to speech recognition \cite{Sak2014LongSM}.

All of these ideas have already been applied to visual question answering task. In fact the model that we describe in this work is very similar to stacked attention networks \cite{Yang2016StackedAN}, nevertheless we show significant improvement over their result ($5.8\%$ on VQA 1.0 dataset). While more recently much more complex and expensive attention models have been explored \cite{Fukui2016MultimodalCB,Nam2016DualAN,Lu2016HierarchicalQC} their advantage is unclear in the light of the results reported in this paper. 

\section{Method}

Figure \ref{fig:overview} shows an overview of our model. In this section we formalize the problem and explain our approach in more detail.

We treat visual question answering task as a classification problem. Given an image $I$ and a question $q$ in the form of natural language we want to estimate the most likely answer $\hat{a}$ from a fixed set of answers  based on the content of the image.
\begin{equation}
\hat{a} = \argmax_a P(a|I, q)
\end{equation}
where $a \in \{a_1, a_2, ..., a_M\}$. The answers are chosen to be the most frequent answers from the training set.

\subsection{Image embedding}
We use a pretrained convolutional neural network (CNN) model based on residual network architecture \cite{Krizhevsky2012ImageNetCW} to compute a high level representation $\phi$ of the input image $I$.
\begin{align}
\phi &= \CNN(I)
\end{align}
$\phi$ is a three dimensional tensor from the last layer of the residual network \cite{He2016DeepRL} before the final pooling layer with $14 \times 14 \times 2048$ dimensions. We furthermore perform $l_2$ normalization on the depth (last) dimension of image features which enhances learning dynamics.

\subsection{Question embedding}
We tokenize and encode a given question $q$ into word embeddings $E_q = \{\mathbf{e}_1, \mathbf{e}_2, ..., \mathbf{e}_P\}$ where $\mathbf{e}_i \in \mathcal{R}^D$, $D$ is the length of the distributed word representation, and $P$ is the number of words in the question. The embeddings are then fed to a long short-term memory (LSTM) \cite{Hochreiter1997LongSM}.
\begin{align}
\mathbf{s} &= \LSTM(E_q)
\end{align}
We use the final state of the LSTM to represent the question. 

\subsection{Stacked  attention}
Similar to \cite{Yang2016StackedAN}, we compute multiple attention distributions over the spatial dimensions of the image features.
\begin{align}
\alpha_{c,l} &\propto \exp F_c(\mathbf{s}, \phi_l)  \quad\ni\quad  \sum_{l=1}^L\alpha_{c, l}=1 \\
\mathbf{x}_c &= \sum_l \alpha_{c, l} \phi_l
\end{align}
Each image feature glimpse $\mathbf{x}_c$ is the weighted average of image features $\phi$ over all the spatial locations $l=\{1, 2, ..., L\}$. The attention weights $\alpha_{c, l}$ are normalized separately for each glimpse $c = 1, 2,..., C$. 

In practice $F = [F_1, F_2, ..., F_C]$ is modeled with two layers of convolution. Consequently $F_i$'s share parameters in the first layer.
We solely rely on different initializations to produce diverse attention distributions. 

\subsection{Classifier}
Finally we concatenate the image glimpses along with the LSTM state and apply nonlinearities to produce probabilities over answer classes.
\begin{align}
P(a_i|I, q) &\propto \exp G_i(\mathbf{x}, \mathbf{s})
\end{align}
where 
\begin{align}
\mathbf{x} &= [\mathbf{x}_1, \mathbf{x}_2, ..., \mathbf{x}_C].
\end{align}
$G = [G_1, G_2, ..., G_M]$ in practice is modeled with two fully connected layers.

Our final loss is defined as follows.
\begin{equation}
\mathcal{L} = \frac{1}{K}\sum_{k=1}^K - \log P(a_k|I, q)
\end{equation}
Note that we average the log-likelihoods over all the correct answers $a_1, a_2, ..., a_K$.

\section{Experiments}
\subsection{Dataset}
\subsubsection{VQA 1.0}
We evaluate our model on both balanced and unbalanced versions of VQA dataset.
VQA 1.0 \cite{Antol2015VQAVQ} is consisted of 204,721 images form the MS COCO dataset \cite{Lin2014MicrosoftCC}.
We evaluate our models on the real open ended challenge which consists of  614,163 questions and 6,141,630 answers. The dataset comes with predefined train, validation, and test splits. There is also a 25\% subset of the the test set which is referred to as test-dev split. For most of experiments we used the train set as training data and reported the results on the validation set. To be comparable to prior work we additionally train our default model on train and val set and report the results on test set.

\subsubsection{VQA 2.0}
We also evaluate our model on the more recent VQA 2.0 \cite{Goyal2016MakingTV} which is consisted of 658,111 questions and 6,581,110 answers. This version of the dataset is more balanced in comparison to VQA 1.0. Specifically for every question there are two images in the dataset that result in two different answers to the question. At this point only the train and validation sets are available. We report the results on validation set after training on train set.

\subsection{Evaluation metric}
We evaluate our models on the open ended task of VQA challenge with the provided accuracy metric.
\begin{equation}
Acc(a) = \frac{1}{K} \sum_{k = 1}^{K} \min(\frac{\sum_{1 \leq j \leq K , j \neq k}\mathbbm{1}(a = a_j)}{3}, 1)
\label{eq:accuracy}
\end{equation}
where $a_1, a_2, ..., a_K$ are the correct answers provided by the user and $K = 10$. Intuitively, we consider an answer correct if at least three annotators agree on the answer. To get some level of robustness we compute the accuracy over all 10 choose 9 subsets of ground truth answers and average.

\section{Results}

\begin{table*}
	\footnotesize
	\centering
	\tabcolsep=0.5cm
	\begin{tabular}{l c c c c c c c c}
		\toprule
		Steps &  1K & 3K & 6K & 12K & 25K & 50K & 100K & 200K \\
		\midrule
		Default & 37.16 & 46.96 & 55.07 & 58.12 & 59.76 & 60.65 & 60.94 & 60.95 \\
		No $l_2$ normalization & 42.65 & 44.87 & 49.07 & 51.12 & 51.75 & 52.15 & 53.56 & 54.69 \\
		No dropout on FC/Conv layers & 32.78 & 38.19 & 49.02 & 58.63 & 57.90 & 57.42 & 57.30 & 56.98 \\
		No dropout on LSTM layers & 45.85 & 51.55 & 55.75 & 57.63 & 59.60 & 59.79 & 59.95 & 59.80 \\
		No attention & 38.09 & 48.36 & 51.42 & 54.43 & 56.02 & 57.13 & 57.65 & 57.72 \\
		Sampling loss & 47.24 & 47.67 & 51.80 & 54.85 & 56.69 & 57.62 & 58.85 & 59.44 \\
		With positional features & 33.26 & 41.37 & 55.36 & 57.95 & 59.75 & 60.44 & 61.02 & 61.09 \\
		Bidirectional LSTM & 42.33 & 52.38 & 55.93 & 58.32 & 59.99 & 60.63 & 60.69 & 60.63 \\
		\midrule
		Word embedding size: 100 & 39.53 & 50.24 & 53.94 & 56.74 & 58.92 & 59.96 & 60.75 & 60.90 \\
		Word embedding size: 300 (default) & 37.16 & 46.96 & 55.07 & 58.12 & 59.76 & 60.65 & 60.94 & 60.95 \\
		Word embedding size: 500 & 37.21 & 47.15 & 55.44 & 58.43 & 59.98 & 60.60 & 61.01 & 61.04 \\
		\midrule
		LSTM state size: 512 & 46.59 & 51.20 & 55.33 & 57.96 & 59.46 & 60.31 & 60.79 & 61.09 \\
		LSTM state size: 1024 (default) & 37.16 & 46.96 & 55.07 & 58.12 & 59.76 & 60.65 & 60.94 & 60.95 \\
		LSTM state size: 2048 & 33.24 & 39.11 & 50.86 & 57.48 & 59.75 & 60.65 & 60.93 & 60.80 \\
		LSTM state size: 1024 1024 & 37.78 & 48.19 & 54.28 & 57.20 & 59.34 & 60.22 & 60.62 & 60.75 \\
		\midrule
		Attention size: 512 1 & 36.54 & 45.74 & 54.23 & 57.42 & 59.46 & 60.22 & 60.85 & 60.96 \\
		Attention size: 512 2 (default) & 37.16 & 46.96 & 55.07 & 58.12 & 59.76 & 60.65 & 60.94 & 60.95 \\
		Attention size: 512 3 & 36.26 & 45.16 & 55.22 & 57.96 & 59.77 & 60.60 & 60.87 & 61.12 \\
		Attention size: 1024 1 & 45.60 & 50.72 & 54.61 & 57.57 & 59.52 & 60.46 & 60.92 & 60.92 \\
		Attention size: 1024 2 & 35.04 & 42.72 & 55.56 & 58.03 & 59.66 & 60.54 & 61.14 & 61.10 \\
		\midrule
		Classifier size: 3000 & 30.19 & 43.12 & 53.38 & 56.18 & 57.82 & 58.25 & 58.24 & 58.12 \\
		Classifier size: 1024 3000 (default) & 37.16 & 46.96 & 55.07 & 58.12 & 59.76 & 60.65 & 60.94 & 60.95 \\
		Classifier size: 2048 3000 & 48.28 & 52.57 & 56.02 & 58.51 & 59.96 & 60.46 & 60.84 & 60.95 \\
		Classifier size: 1024 1024 3000 & 44.51 & 49.53 & 53.25 & 55.95 & 57.59 & 58.83 & 60.05 & 60.66 \\
		\bottomrule
	\end{tabular}
	\vspace{.1cm}
	\caption{This table shows the result of different mutations of our default model. All models are trained on training set of VQA 1.0 \cite{Antol2015VQAVQ} and the accuracy is reported on validation set according to equation \ref{eq:accuracy}.  Applying $l_2$ normalization, dropout, and using soft-attention significantly improves the accuracy of the model. Some of the previous works such as \cite{Fukui2016MultimodalCB} had used the sampling loss, which we found to be leading to significantly worse results and longer training time. Different word embedding sizes and LSTM configurations were explored but we found it to be not a major factor.  Contrary to results reported by \cite{Yang2016StackedAN} we found using stacked attentions to only marginally improve the result. We found a two layer deep classifier to be significantly better than a single layer, adding more layers or increasing the width did not seem to improve the results.} 
	\label{table:baselines}
\end{table*}

\subsection{Baselines}
In this section we describe the details of our default baseline as well as its mutations.

In all of the baselines input images are scaled while preserving aspect ratio and center cropped to $299 \times 299$ dimensions. We found stretching the image to harm the performance of the model. Image features are extracted from pretrained 152 layer ResNet \cite{He2016DeepRL} model. We take the last layer before the average pooling layer (of size $14 \times 14 \times 2048$) and perform $l_2$ normalization in the depth dimension. 

The input question is tokenized and embedded to a $D = 300$ dimensional vector. The embeddings are passed through $\tanh$ nonlinearity before feeding to the LSTM. The state size of LSTM layer is set to $1024$. Per example dynamic unrolling is used to allow for questions of different length, although we cap maximum length of the questions at $15$ words.

To compute attention over image features, we concatenate tiled LSTM state with image features over the depth dimension and pass through a $1 \times 1$ dimensional convolution layer of depth $512$ followed by ReLU \cite{Nair2010RectifiedLU} nonlinearity. The output feature is passed through another $1 \times 1$ convolution of depth $C = 2$ followed by softmax over spatial dimensions to compute attention distributions. We use these distributions to compute two image glimpses by computing the weighted average of image features.

We further concatenate the image glimpses with the state of the LSTM and pass through a fully connected layer of size $1024$ with ReLU nonlinearity. The output is fed to a linear layer of size $M = 3000$ followed by softmax to produce probabilities over most frequent classes.

We only consider top $M = 3000$ most frequent answers in our classifier. Other answers are ignored and do not contribute to the loss during training. This covers $92\%$ of the answers in the validation set in VQA dataset \cite{Antol2015VQAVQ}.

We use dropout of $0.5$ on input features of all layers including the LSTM, convolutions, and fully connected layers.

We optimize this model with Adam optimizer \cite{Kingma2014AdamAM} for $100K$ steps with batch size of $128$. We use exponential decay to gradually decrease the learning rate according to the following equation.
$$
l_{step} = 0.5^{\frac{\text{step}}{\text{decay steps}}} l_0
$$
The initial learning rate is set to $l_0 = 0.001$, and the decay steps is set to $50K$. We set $\beta_1 = 0.9$ and $\beta_2 = 0.999$.

During training CNN parameters are kept fixed. The rest of the parameters are initialized as suggested by Glorot et al. \cite{Glorot2010UnderstandingTD}.

Table \ref{table:baselines} shows the performance of different baselines on validation set of VQA 1.0 \cite{Antol2015VQAVQ} when trained on the training set only. We have reported results for the following mutations of our default model:
\begin{itemize}
	\item \textbf{No $l_2$ norm}: ResNet features are not $l_2$ normalized.
	\item \textbf{No dropout on FC/Conv}: Dropout is not applied to the inputs of fully connected and convolution layers.
	\item \textbf{No dropout on LSTM}: Dropout is not applied to the inputs of LSTM layers.
	\item \textbf{No attention}: Instead of using soft-attention we perform average spatial pooling before feeding image features to the classifier.
	\item \textbf{Sampled loss}: Instead of averaging the log-likelihood of correct answers we sample one answer at a time.	
	\item \textbf{With positional features}: Image features $\phi$ are augmented with $x$ and $y$ coordinates of each cell along the depth dimension producing a tensor of size $14 \times 14 \times 2050$.
	\item \textbf{Bidirectional LSTM}: We use a bidirectional LSTM to encode the question.	
	\item \textbf{Word embedding size}: We try word embeddings of different sizes including $100$, $300$ (default), and $500$.
	\item \textbf{LSTM state size}: We explore different configurations of LSTM state sizes, this include a one layer LSTM of size $512$, $1024$ (default), and $2048$ or a stacked two layer LSTM of size $1024$.
	\item \textbf{Attention size}: Different attention configurations are explored. First number indicates the size of first convolution layer and the second number indicates the number of attention glimpses.
	\item \textbf{Classifier size}: By default classifier $G$ is consisted of a fully connected layer of size $1024$ with ReLU nonlinearity followed by a $M = 3000$ dimensional linear layer followed by softmax. We explore shallower, deeper, and wider alternatives.
\end{itemize}

$l_2$ normalization of image features improved learning dynamics leading to significantly better accuracy while reducing the training time.

We observed that applying dropout on multiple layers (including fully connected layers, convolutions, and LSTMs) is crucial to avoid over-fitting on this dataset.

As widely reported we confirm that using soft-attention significantly improves the accuracy of the model. 

Different word embedding sizes and LSTM configurations were explored but we found it to be not a major factor. A larger embedding size with a smaller LSTM seemed to work best.

Some of the previous works such as \cite{Fukui2016MultimodalCB} had used the sampling loss, which we found to be leading to significantly worse results and longer training time.

Contrary to results reported by \cite{Yang2016StackedAN} we found using stacked attentions to only marginally improve the result.

We found a two layer deep classifier to be significantly better than a single layer, adding more layers or increasing the width did not seem to improve the results. 

\begin{table*}
	\footnotesize
	\centering
	\tabcolsep=0.6cm
	\begin{tabular}{l c c c c c c c c}
		\toprule
		\multirow{4}{*}{\raisebox{-\heavyrulewidth}{Method}} & 
		
		\multicolumn{4}{c}{Test-Dev} & 
		\multicolumn{4}{c}{Test-Standard} \\
		\cmidrule(lr){2-5}
		\cmidrule(lr){6-9}
		& Y/N & Num & Other & All  & Y/N & Num & Other  & All \\ 
		\midrule[1pt]
		VQA team \cite{Antol2015VQAVQ} & 80.5 & 36.8 & 43.1 & 57.8 & 80.6 & 36.5 & 43.7 & 58.2 \\
		SAN (VGG) \cite{Yang2016StackedAN} & 79.3 & 36.6 & 46.1 & 58.7 & - & - & - & 58.9 \\
		NMN (VGG) \cite{Andreas2016NeuralMN} & 81.2 & 38.0 & 44.0 & 58.6 & - & - & - & 58.7 \\
		ACK (VGG) \cite{Wu2016AskMA} & 81.0 & 38.4 & 45.2 & 59.2 & 81.1 & 37.1 & 45.8 & 59.4 \\
		DMN+ (VGG) \cite{Xiong2016DynamicMN} & 80.5 & 36.8 & 48.3 & 60.3 & - & - & - & 60.4 \\
		MRN (ResNet) \cite{Kim2016MultimodalRL} & 82.3 & 38.8 & 49.3 & 61.7 & 82.4 & 38.2 & 49.4 & 61.8 \\
		HieCoAtt (ResNet) \cite{Lu2016HierarchicalQC} & 79.7 & 38.7 & 51.7 & 61.8 & - & - & - & 62.1 \\
		DAN (VGG) \cite{Nam2016DualAN} &  82.1 & 38.2 & 50.2 & 62.0 & -  &  - &  - & - \\
		RAU (ResNet) \cite{Noh2016TrainingRA} & 81.9 & 39.0 & 53.0 & 63.3 & 81.7 & 38.2 & 52.8 & 63.2 \\
		MCB (ResNet) \cite{Fukui2016MultimodalCB} & 82.2 & 37.7 & 54.8 & 64.2 & - & - & - & - \\
		DAN (ResNet) \cite{Nam2016DualAN} & \textbf{83.0} & \textbf{39.1} & 53.9 & 64.3 & \textbf{82.8} & 38.1 & 54.0 & 64.2 \\
		\midrule
		Ours (ResNet) & 82.2 & \textbf{39.1} & \textbf{55.2} & \textbf{64.5} & 82.0 & \textbf{39.1} & \textbf{55.2} & \textbf{64.6} \\
		\bottomrule
	\end{tabular}
	\vspace{.1cm}
	\caption{This table shows a comparison of our model with state of the art on VQA 1.0 dataset. While our model is architecturally simpler and smaller in terms of trainable parameters than most existing work, nevertheless it outperforms all the previous work.} 
	\label{table:comp_vqa1}
\end{table*}

\begin{table}
	\footnotesize
	\centering
	\tabcolsep=0.4cm
	\begin{tabular}{l c c c c}
		\toprule
		Method & Y/N & Num & Other & All \\ 
		\midrule
		HieCoAtt \cite{Lu2016HierarchicalQC} & 71.80 & 36.53 & 46.25 & 54.57 \\
		MCB \cite{Fukui2016MultimodalCB} & 77.37 & 36.66 & 51.23 & 59.14 \\
		\midrule
		Ours & \textbf{77.45} & \textbf{38.46} & \textbf{51.76} & \textbf{59.67} \\
		\bottomrule
	\end{tabular}
	\vspace{.1cm}
	\caption{Our results on VQA 2.0 \cite{Goyal2016MakingTV} validation set when trained on the training set only. Our model achieves an overall accuracy of $59.67\%$ which marginally outperforms state of the art on this dataset.} 
	\label{table:comp_vqa2}
\end{table}

\subsection{Comparison to state of the art}
Table \ref{table:comp_vqa1} shows the performance of our model on VQA 1.0 dataset. We trained our model on train and validation set and tested the performance on test-standard set. Our model achieves an overall accuracy of $64.6\%$ on the test-standard set, outperforming best previously reported results by $0.4\%$. All the parameters here are the same as the default model. 

While architecturally our default model is almost identical to \cite{Yang2016StackedAN}, some details are different. For example they use the VGG \cite{Simonyan2014Very} model, while we use ResNet \cite{He2016DeepRL} to compute image features. They do not mention $l_2$ normalization of image features which found to be crucial to reducing training time.
They use SGD optimizer with momentum $\mu = 0.9$, while we found that Adam \cite{Kingma2014AdamAM} generally leads to faster convergence. 

We also reported our results on VQA 2.0 dataset \ref{table:comp_vqa2}. At this point we only have access to train and validation splits for this dataset. So we trained the same model on the training set and evaluated the model on the validation set. Overall our model achieves $59.67\%$ accuracy on the validation set which is about $0.5\%$ higher than best previously reported results.

\begin{figure*}
	\centering
	\subfigure[What brand is the shirt?]{
		\includegraphics[height=80pt]{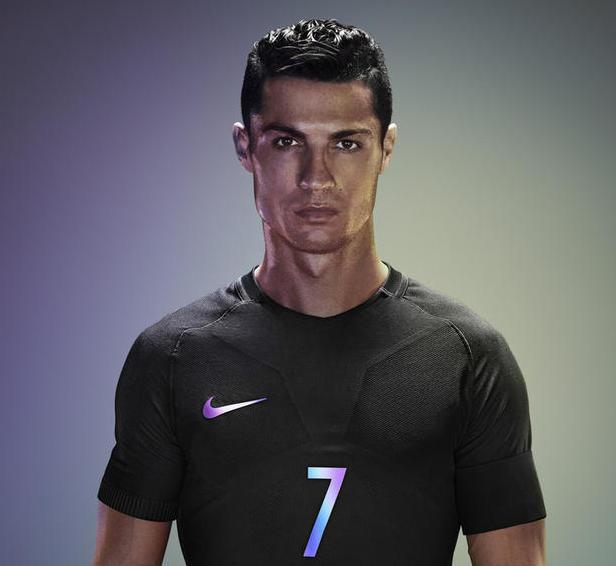}
		\includegraphics[height=80pt]{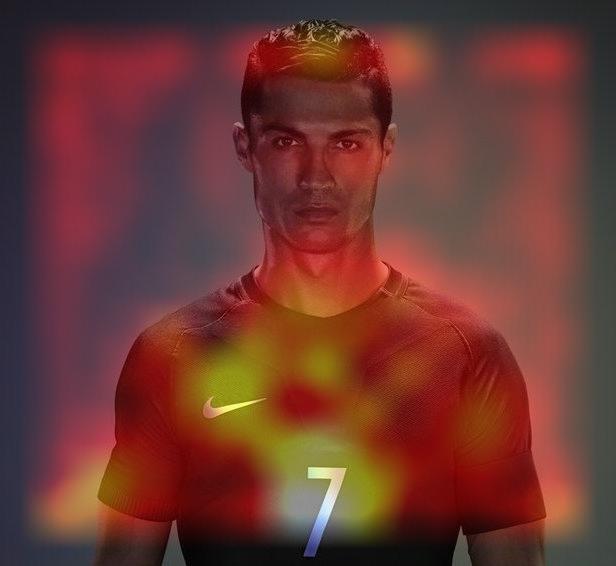}
		\includegraphics[height=80pt]{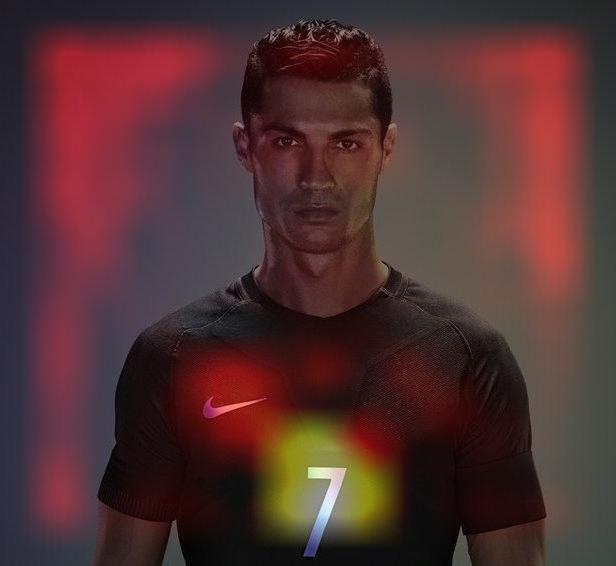}
		\includegraphics[height=50pt]{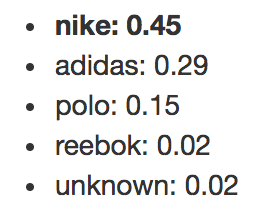}
	}
	\subfigure[What time is it?]{
		\includegraphics[height=80pt]{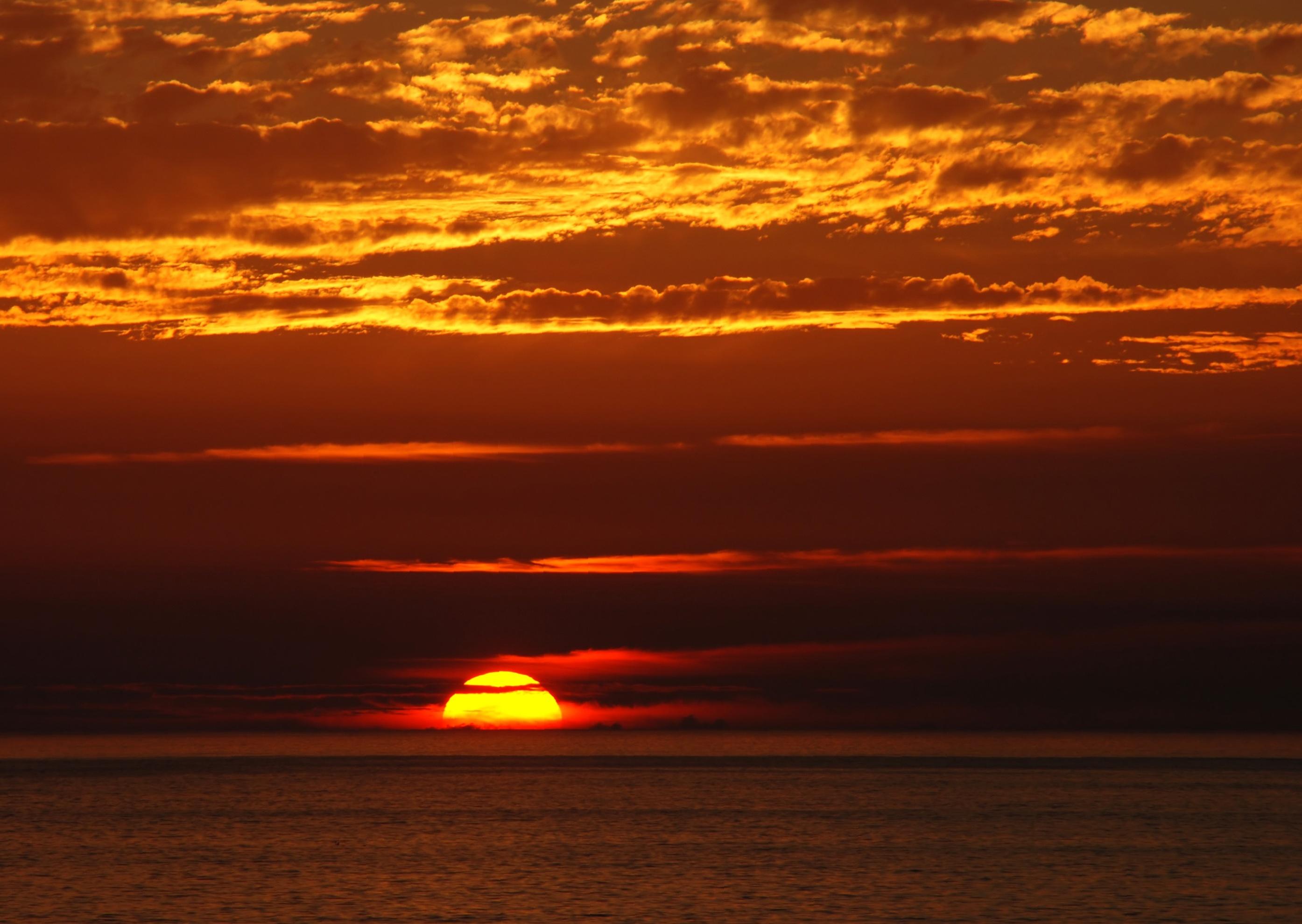}
		\includegraphics[height=80pt]{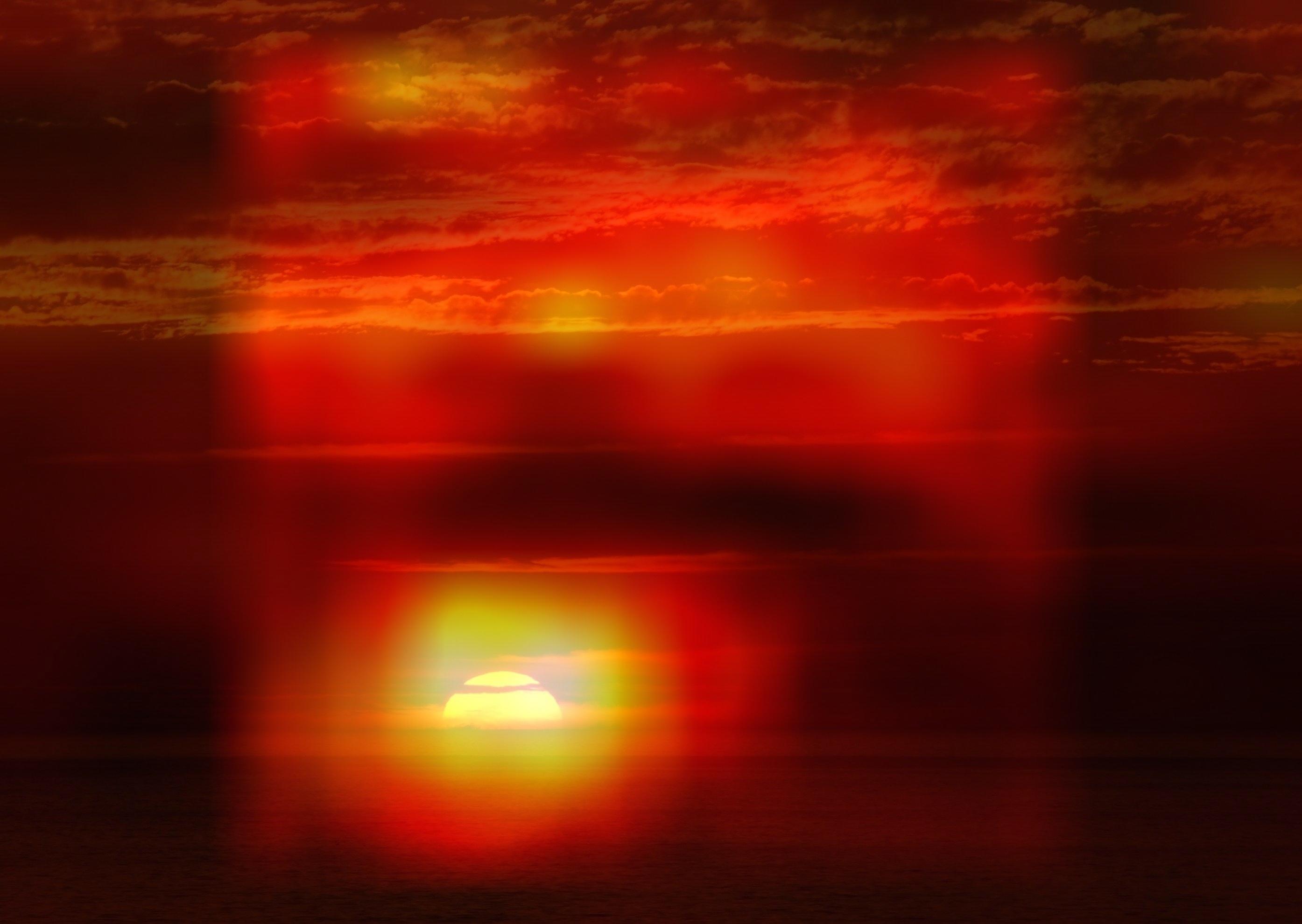}
		\includegraphics[height=80pt]{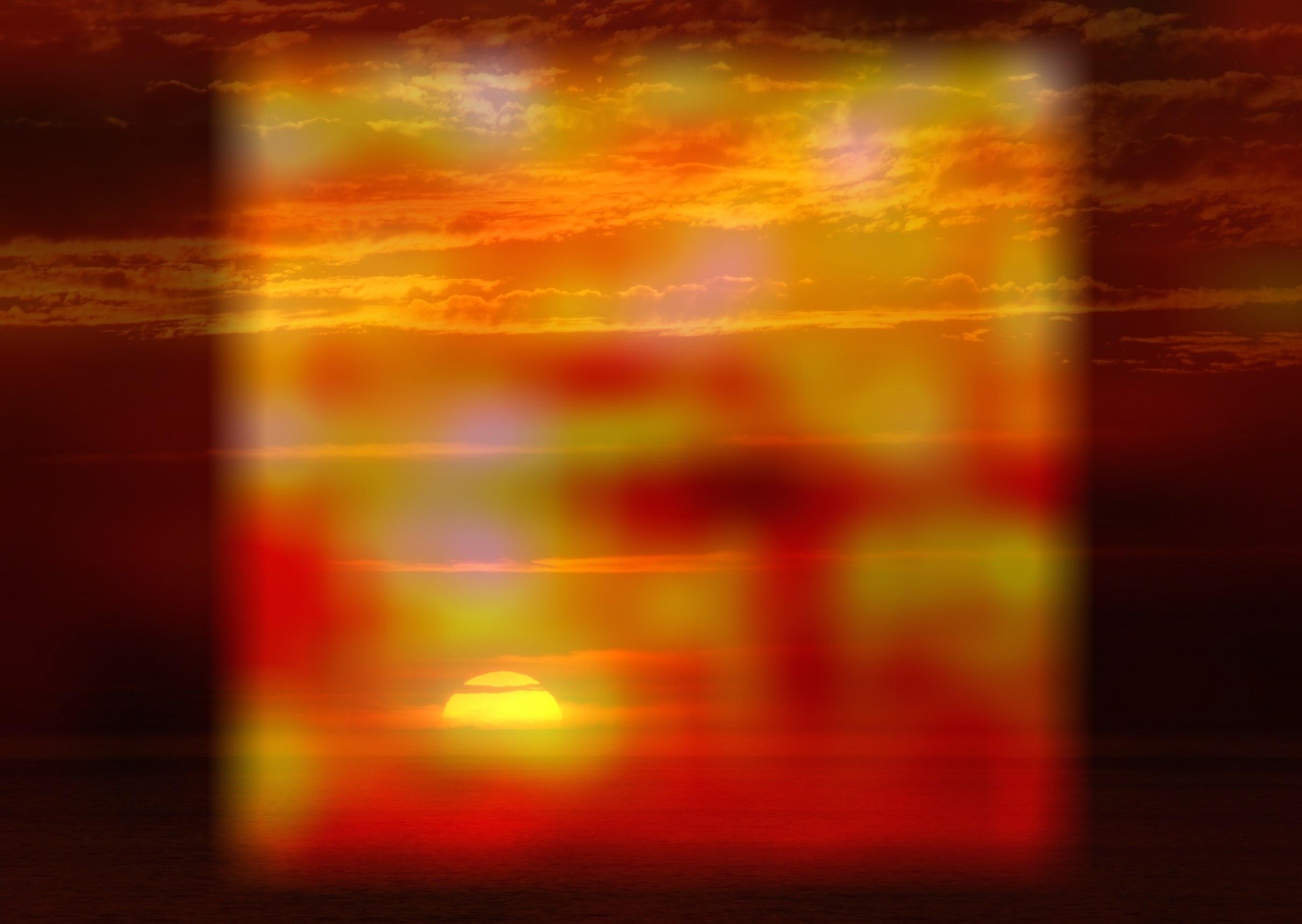}
		\includegraphics[height=50pt]{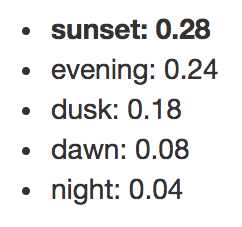}
	}
	\subfigure[How does the man feel?]{
		\includegraphics[height=80pt]{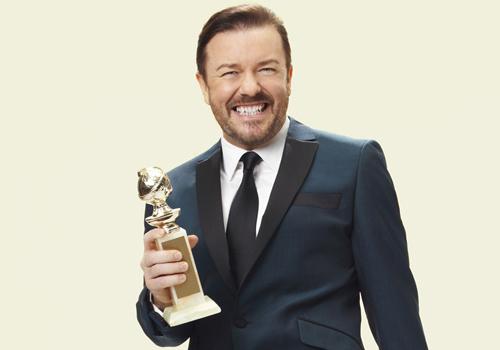}
		\includegraphics[height=80pt]{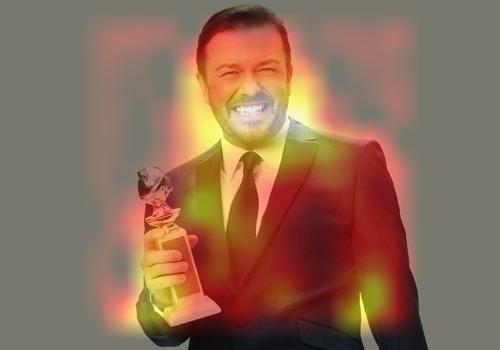}
		\includegraphics[height=80pt]{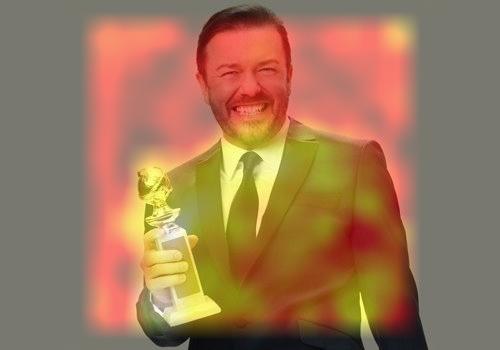}
		\includegraphics[height=50pt]{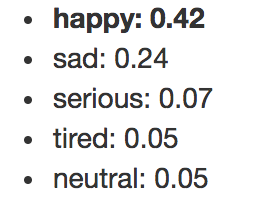}
	}
	\subfigure[What is the girl doing?]{
		\includegraphics[height=80pt]{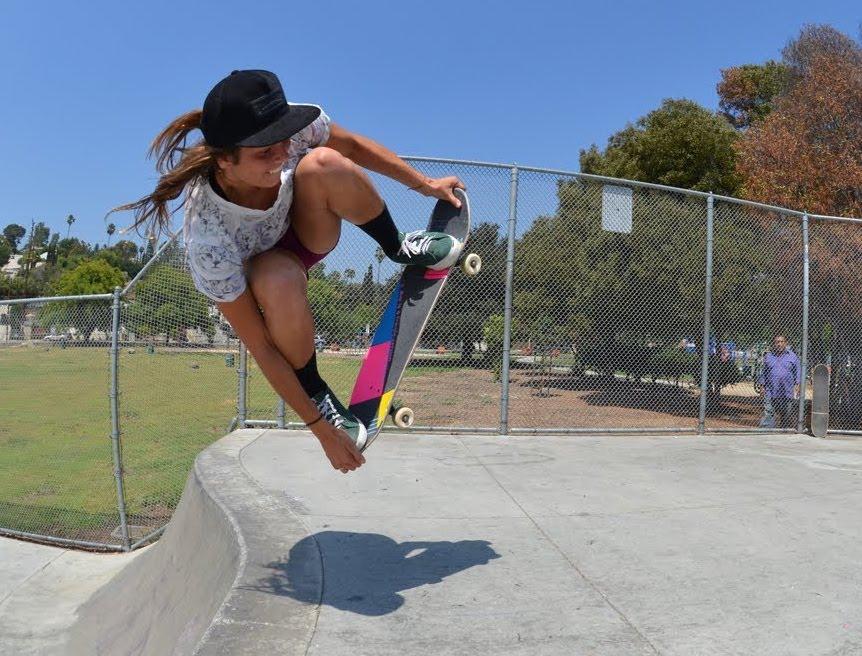}
		\includegraphics[height=80pt]{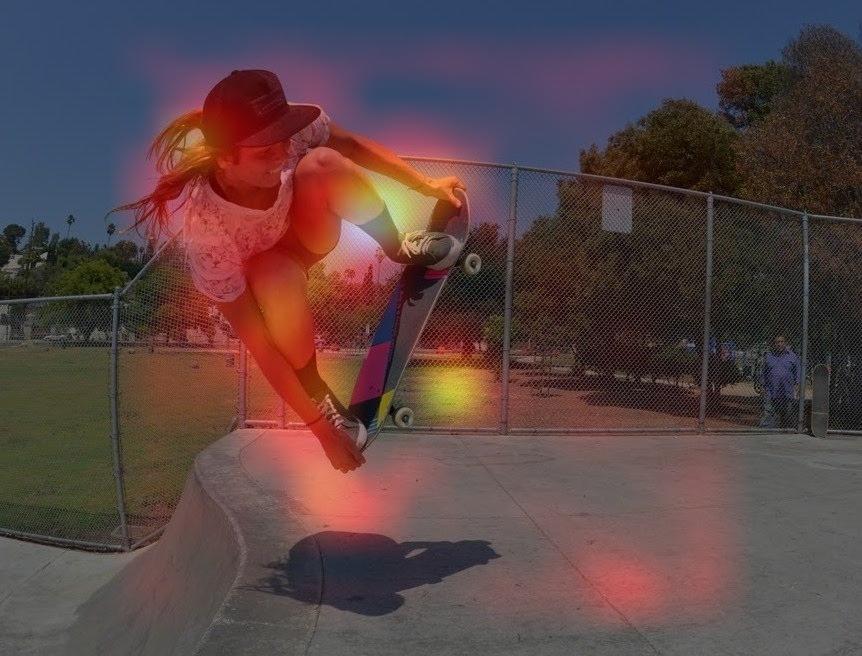}
		\includegraphics[height=80pt]{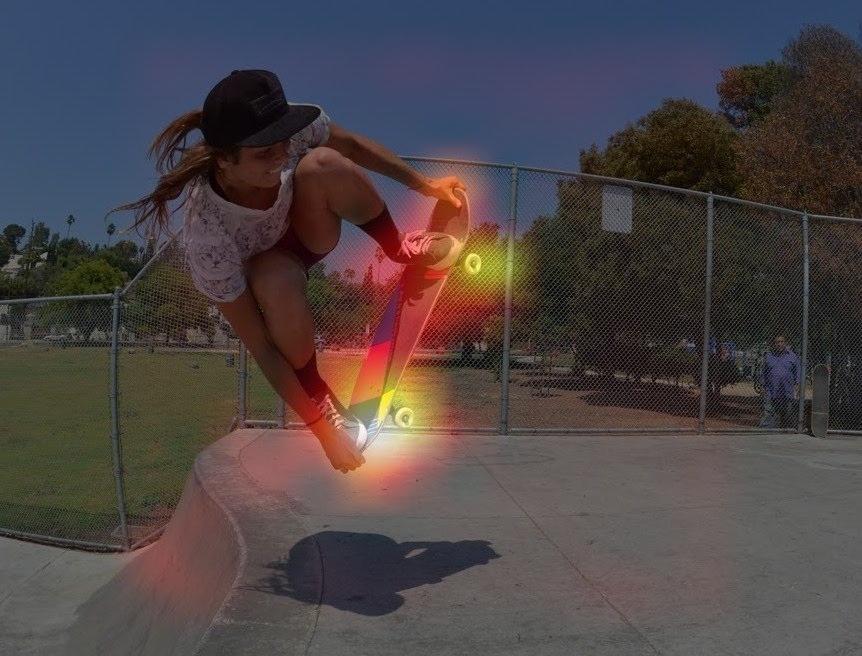}
		\includegraphics[height=50pt]{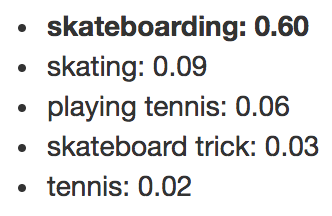}
	}
	\caption{Qualitative results on sample images shows that our model can produce reasonable answers to a range of questions.}
	\label{fig:image1}
\end{figure*}

\section{Conclusion}
In this paper we presented a new baseline for visual question answering task that outperforms previously reported results on VQA 1.0 and VQA 2.0 datasets. Our model is architecturally very simple and in essence very similar to the models that were tried before, nevertheless we show once the details are done right this model outperforms all the previously reported results.

{\small
\bibliographystyle{ieee}
\bibliography{egbib}
}

\end{document}